\documentclass[sigconf]{acmart} 
\AtBeginDocument{%
  }

\setcopyright{acmlicensed}
\copyrightyear{2018}
\acmYear{2018}
\acmDOI{XXXXXXX.XXXXXXX}
\acmConference[Conference acronym 'XX]{Make sure to enter the correct
  conference title from your rights confirmation email}{June 03--05,
  2018}{Woodstock, NY}

\acmISBN{978-1-4503-XXXX-X/2018/06}

\usepackage{algorithmic}
\usepackage{algorithm2e}
\usepackage{dblfloatfix}
\usepackage{cuted}
\usepackage{caption}

\begin{document}

\title{An Exploration of Collision-based Enemy Morphology Generation}



\author{
    Johor Jara Gonzalez \quad Matthew Guzdial \\
    Alberta Machine Intelligence Institute (Amii)\\
    Department of Computing Science, \\
    University of Alberta \\
    \texttt{\{jaragonz, guzdial\}@ualberta.ca}
}


\begin{abstract}
Despite a great deal of prior research into Procedural Content Generation (PCG), relatively little prior work has explored generating enemies for video games. 
In particular, there is almost no work on generating enemy morphologies, the basic body plan or collision information for in-game enemies, despite the existence of related morphology generation work in robotics. 
In this paper, we explore three different novel approaches to generate enemy morphologies based on player collision information.
We found that each approach provides different strengths and weaknesses, but all had equivalent or better performance than an evolutionary baseline adapted from prior robotics morphology work. 

\end{abstract}



\begin{teaserfigure}
  \centering
  \includegraphics[width=\textwidth]{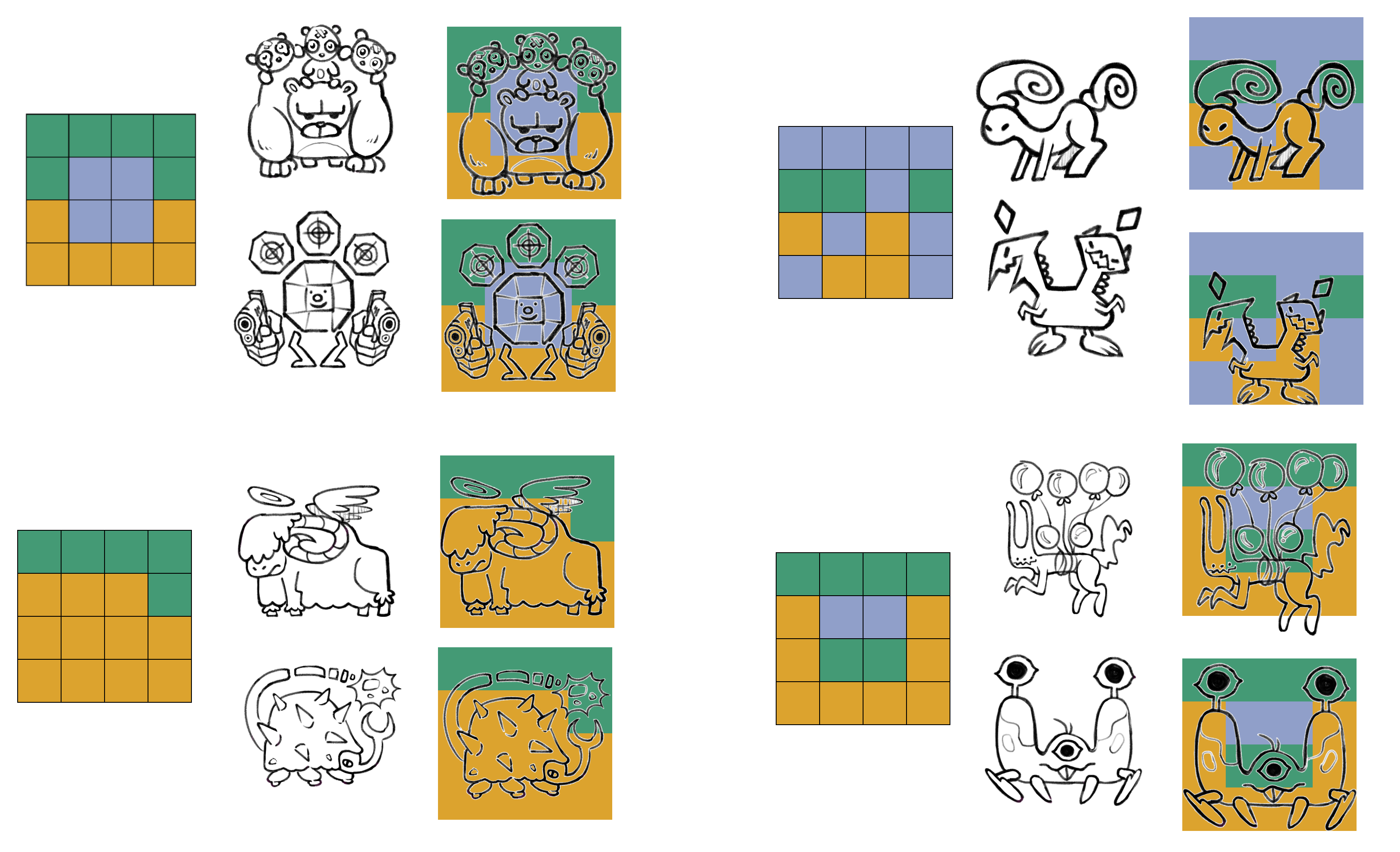}
  \caption{Output of the enemy morphology representation from different generators along with artist concepts using these as inspiration for ideation. We include overlays of the image made by the artist and the output of the generator. }
  \label{fig:teaser}
\end{teaserfigure}

\maketitle

\section{Introduction}
Procedural Content Generation(PCG) refers to the use of algorithmic techniques to automatically create game content such as levels, maps, and characters rather than authoring assets by hand \cite{zhang2022survey,hendrikx2013procedural,guzdial2025procedural}. 
Within this broad area, enemy generation is concerned with automatically constructing the adversaries players face in combat in games, including their behaviours, statistics, and spatial configurations. In this paper, we focus on enemy morphology: the body plan and collision information that determine the direct interactions between an enemy and a player character. Although morphology generation has been extensively explored in robotics and virtual creatures where evolutionary and learning-based methods construct bodies and controllers simultaneously \cite{song2024morphvae,ha2019reinforcement,lehman2011evolving,sims2023evolving}, to our knowledge there is almost no work in games for enemy morphology generation, especially not in a way that is explicitly conditioned on how those enemies interact with the player.

A natural starting point for our work might be to adapt morphology generation methods from robotics to games. However, these methods are formulated around a different optimization problem. In evolutionary robotics and virtual creatures, morphology is optimized for the agent itself, bodies are evolved to maximize locomotion quality, stability, or task performance)\cite{song2024morphvae,ha2019reinforcement,lehman2011evolving,sims2023evolving}. The ``player'' in these systems is the evolving agent, and evaluation is typically based on how well it moves or survives. By contrast, game enemies are designed to shape the experience of a separate player agent whose features are fixed by the game design. Our goal is not to maximize an enemy's own performance, but to structure how a player must act in order to defeat it. For example, robotics work does not usually optimize a robot's morphology to make it easier (or harder) for another robot to destroy it: in games, this relationship between player and enemy is central. All together this makes a direct transfer of these approaches inadequate.

In the absence of morphology generation techniques, enemy generation has instead focused on other aspects: behaviour generation, \cite{gutierrez2021reinforcement, zook2014automatic, halina2022diversity,ponton2025analyzing} optimizing enemy features   \cite{pereira2021procedural, cabrera2015procedural}, and mechanics \cite{zook2014generating}. The closest work to ours still does not generate new enemy morphologies.  Butler et. al. structure boos encounters and attack patterns, but assume fixed, designer-authored morphologies\cite{butler2017program}. Cook and Colon's A Rogue Dream generates new sprites that reskin existing entities without changing their collision information \cite{cook2014rogue}. As a result, most works operate over fixed enemy morphologies- varying behaviours, parameters, and appearance, but not the morphology itself.

To motivate our approach, consider a hypothetical game partway through development: a metroidvania-style platformer with several movement mechanics already implemented (e.g., wall jump, air dash, etc.). At this stage designers may want enemies that can only be defeated using specific mechanics - both to teach those mechanics and to gate access to new regions of the world. Designers must explore a large space of possible enemy morphologies and placements, while ensuring that the target mechanic can reliably defeat the enemy in realistic situations and other mechanics cannot easily bypass or trivialize the encounter. A system that could propose enemy morphologies satisfying constraints such as ``defeatable by mechanic A but not mechanic B'', based on observed interaction data rather than hand-specified fitness function or tedious playtesting, could substantially reduce this burden and give designers a new axis of expressive control by generating constraint-satisfying enemy proposals that serve as concrete starting points for iteration.

In this paper, we treat enemy generation as a morphology design problem driven by player-enemy interaction data. We develop a grid-base 2D environment for enemy generation in which enemies are represented by their collision information on a discrete grid, where player agents with different mechanics interact with them. We explore three optimization strategies based on reinforcement learning, A* search, and neural networks which learn from player-enemy interaction data.   
To contextualize these methods, we implement an evolutionary algorithm baseline inspired by morphology generation work in robotics and virtual creatures\cite{song2024morphvae}. We evaluate all approaches on a case study where the design goal is to generate enemies that are defeatable by one set of mechanic but not another, using generated mechanics drawn from prior research. Across out experiments, we find that all three interaction-driven approaches achieve comparable or better performance than the evolutionary baseline, while exhibiting distinct trade-offs.

Our contributions on this paper are the following: 

\begin{itemize}
    \item A grid-based enemy generation environment that represents enemy morphologies in terms of collision information, and records detailed player-enemy interaction traces  \footnote{https://github.com/Harcurio/Enemy-Morphology-Generation}.
    \item Three interaction-driven approaches to enemy morphology generation, based on Reinforcement Learning and Neural Networks, A* search, and Reinforcement Learning alone.
    \item An evolutionary morphology-generation baseline adapted from robotics morphology generation research to our problem.
    \item An empirical evaluation demonstrating that our approaches can generate enemies that gate specific mechanics (defeatable by one mechanic but not another), and an analysis of their respective strengths and weaknesses.
\end{itemize}

\section{ Related Work}



Procedural Content Generation (PCG ) refers to the automatic creation of game content using algorithms, such as levels, items, or maps\cite{zhang2022survey,hendrikx2013procedural,guzdial2025procedural}. In this work we are interested in a specific instance of PCG: enemy generation, in particular their morphology. For this, we focus this section on two strands: prior PCG research on enemy generation and morphology generation in robotics, which are most relevant to our work. 

\subsection{Enemy Generation}

Surveys, papers, and textbooks on PCG  tend to prioritize levels, rules, items, stories, and quests , with enemies usually treated peripherally as part of broader discussions of NPCs difficulty balancing or encounter design\cite{zhang2022survey,hendrikx2013procedural,guzdial2025procedural}. Consequently, existing research on enemy generation is fragmented across three largely separate strands: behaviour (how enemies act and adapt), features (their numerical parameters), and mechanics (the rules relevant to them and their in-game abilities).

For behaviour, a large body of work treats enemies as learning agents whose policies are optimized with reinforcement learning or, more recently, steered by large language models. RL-based approaches use enemies (or surrogate agents) to explore and adapt to game situations. Gutiérrez-Sánchez et al.\ combine RL with behaviour trees to automatically playtest stealth AI variants and quantify how design changes affect detection and challenge \cite{gutierrez2021reinforcement}. Merrick and Maher’s motivated RL agents learn evolving behavioural patterns for persistent online worlds rather than relying on fixed rule sets \cite{merrick2006motivated}. More applied work in commercial engines, such as Nämerforslund’s Unity ML-Agents adversaries \cite{namerforslund2021machine} and Maurya et al.’s RL-optimized NPCs \cite{maurya2025optimizing}, trains enemies whose navigation and combat policies adapt dynamically to player behaviour and level layout. On the LLM side, researchers are increasingly experimenting with LLMs as high-level controllers or co-designers for NPCs and adversaries, generating context-sensitive actions, dialog, and even code-level behaviours at runtime\cite{gallotta2024large,maleki2024procedural}. Systems such as Jennings and Hartmann’s GROMIT prototype use GPT-4 to synthesize new Unity behaviours on demand from natural-language prompts and semantic scene graphs, effectively generating enemy and object behaviours as code during play \cite{jennings2024towards}. Hassan and Aboulhassan propose an LLM-centric framework that couples player-behaviour analytics with LLM-driven recommendations to adjust narrative beats and adversary types to sustain engagement \cite{hassan2024framework}. These approaches demonstrate how RL and LLM-based techniques can generate enemy behaviours, but they all assume a fixed, designer-authored enemy. In contrast, we use learning and search-based agents primary as a source of interaction traces to drive the generation of new enemy morphologies, rather than as the main object of optimization themselves.

For enemy features, such as health, damage, speed, and weapon loadouts, work typically uses search-based procedural content generation (PCG): evolutionary algorithms, Monte Carlo–style search, and other heuristic search to generate and balance enemy features against fitness functions that model effectiveness or difficulty \cite{togelius2010search,togelius2014procedural,cabrera2015procedural,zohaib2018dynamic, sorochan2022generating, wood2019understanding}.  
For enemy mechanics—new actions, abilities, or unit rule sets that determine how enemies move, attack, coordinate, and respond to players—research leans on program synthesis and symbolic methods, where systems automatically synthesize game mechanics from low-level engine primitives, specifications, or examples, or generate code-level behaviours via search and synthesis. These include Zook \& Riedl’s mechanic generation, Butler et al. program-synthesis-as-PCG approach, and more recent tools like Mechanic Maker \cite{zook2014automatic,butler2017program,sumner2024mechanic,cook2013mechanic}. These approaches create new behaviours, but again assume that the underlying collision-aware morphology of the enemy is fixed by the designer.


\subsection{Morphology Generation in Robotics}

Outside of games, morphology generation has been extensively studied in the fields of robotics and virtual-creatures, where algorithms jointly optimize morphologies and controllers based on task performance \cite{song2024morphvae,ha2019reinforcement,sims2023evolving,lehman2011evolving}. In this setting, morphology is represented as articulated limbs, voxel based structures, or other compositional encodings. Typically evolutionary algorithms update both morphology and control to maximize objectives such as locomotion speed, stability, or task success. Evaluation is conducted from the perspective of the evolving agents itself. A morphology is considered good if it improves the agent's own ability to move or complete tasks in a given environment. 
In comparison, we focus leverage video game-specific player-enemy interaction information to optimize enemy morphologies.


\section{Problem Definition}

\begin{figure}
    \centering
    \includegraphics[width=1\linewidth]{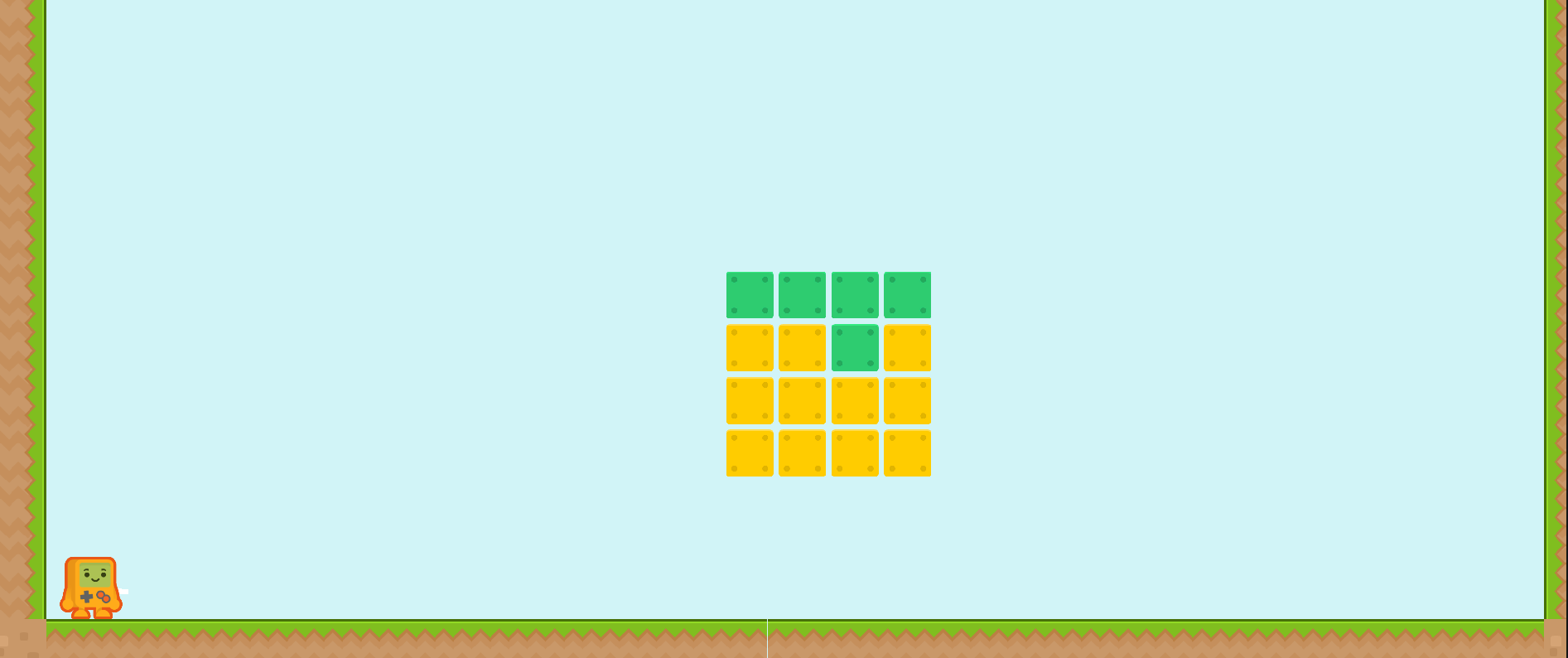}
    \caption{Environment Representation in Unity, where the agent interacts with the grid representing the enemy morphology. The agent will randomly spawn to the left or right of the enemy. }
    \label{fig:placeholder}
\end{figure}
We now formalize the task of generating enemy morphologies that gate specific player mechanics.
We study this problem in a simplified 2D Unity platformer environment designed to collect collision interactions with easy access to Unity's ML Agents while remaining fast to simulate and easy to visualize. This abstraction let us focus on the space of enemy morphologies without confounding factors such as complex level geometry, or  multiple enemies.

Each enemy is instantiated as a fixed 4x4 grid of cells. Each cell in the enemy's grid has one of three types defining a particular interaction on collision with the player:
\begin{itemize}
        \item \textit{weak},  vulnerable cell that can be used to defeat the enemy;
        \item \textit{lethal}, damaging cell that defeats the player agent on contact; and 
        \item \textit{empty} cell, a non-colliding cell.
\end{itemize}

An enemy's morphology is therefore defined as a 4x4 grid over this three-valued encoding, specifying the spatial arrangement of weak, lethal, and empty cells. We denote $E$ as the finite set of all possible 4x4 morphologies, and $e \in E $ as a particular morphology. 

In this environment, the player has a small set of basic mechanics: move left, move right, and jump. We use $B = \{left,right,jump\}$ denote this set of basic mechanics. Our design goal is to generate enemies that are only defeatable when the player agent also has access to an additional mechanic.

To capture this gating scenario, we introduce extra mechanics derived from prior work on automatic mechanic generation \cite{gonzalez2023mechanic}. Let $M$ denote the finite set of extra mechanics (i.e.., vertical teleport, horizontal teleport, double jump, and double speed; see section 4.2 for details) and let $m \in M$ be a particular mechanic. We model each extra mechanic $m$ as an extension to the basic movement mechanics.
For each extra mechanic $m \in M$ we consider two agent configurations:

\begin{itemize}
    \item a baseline configuration  that has access only to the basic mechanics $B$ (left, right, jump); and 
    \item an augmented configuration that has access to the same baseline configuration  plus one of the additional mechanics $B \cup \{m\} $.
\end{itemize}

In different experimental conditions, there configurations are instantiated  either as reinforcement learning agents \cite{juliani2020}, or A* search path finding agents \cite{hart1968formal}. For the purposes of this formulation, we treat them generically as agents that interact with enemies with a specified set of mechanics.

Interactions in the environment are broken into episodes. Each episode begins when an agent is spawned into the environment with a given configuration ($B$ or $B \cup \{m\}$) facing a particular enemy morphology $e \in E$ and ends when either the enemy is defeated, the agent dies, or the episode reaches a fixed time limit. For every episode, we log a time-stamped interaction trace that records the player agent's positions, the enemy cell (weak,lethal) involved in any collision, and the outcome of each collision (enemy destroyed, player agent death). Each trace is associated with the configuration ($B$ or $B \cup \{m\}$) and the underlying enemy morphology $e$. The resulting dataset of traces is stored as JSON files and serves as the data for optimizing the enemy morphology.

Given this setup, our goal is to learn conditional enemy-morphology generators that, for a target extra mechanic $m$, produce 4x4 enemies $e$ with the following properties:

\begin{itemize}
    \item  Defeatability with the mechanic: the agent with the augmented configuration can reliably defeat the enemy
    \item Gating without the mechanic: the agent with a baseline configuration is unable to defeat the same enemy.
\end{itemize}

Formally, we seek a generator $G_\theta(e \mid m)$ that maps a target mechanic $m$ to an enemy morphology $e$
such that the introduced win/loss behaviour of the baseline and augmented agents realizes the desired gating behaviour in the game.

\section{System Overview}
\begin{figure}
    \centering
    \includegraphics[width=1\linewidth]{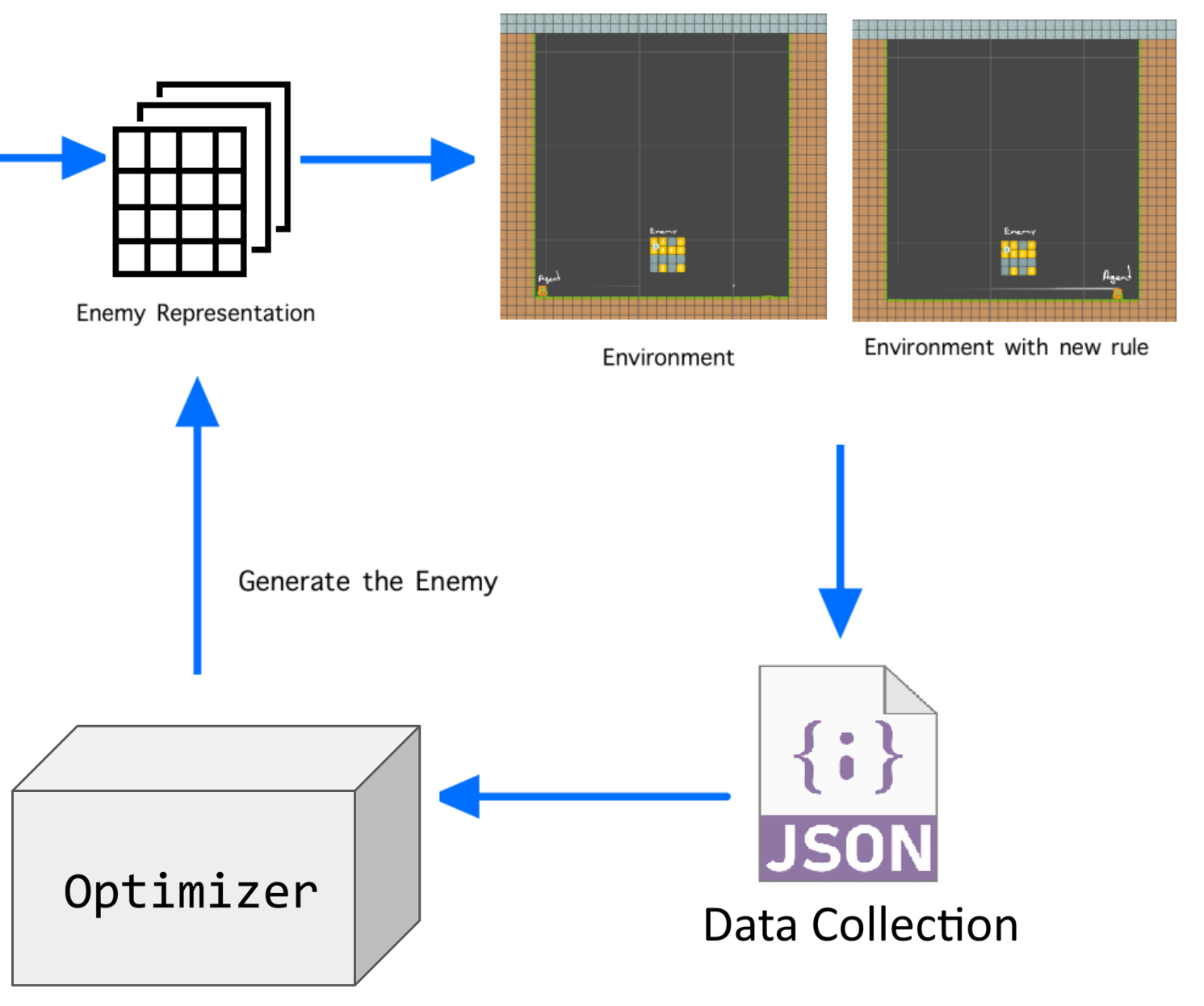}
    \caption{Full pipeline where the first enemy representation are randomly generated}
    \label{fig:fullpipeline}
\end{figure}

The previous section formalized our objective as learning a conditional generator $G_\theta(e \mid m)$ over 4x4 enemy morphologies  $e \in E $, conditioned on a target extra mechanic $m \in M $, such that the resulting enemies exhibit the desired mechanic-gating behaviour. In this section, we describe our specific implementations of this generator. 

At a high level, the pipeline operates as an interaction-driven optimization loop applied to individual enemy morphologies. We begin from a small pool of randomly initialized morphologies that serve as independent starting points. For a given target mechanic $m$, agents using the baseline configuration and the augmented configuration (as defined in Section 3) interact with the selected morphology $e$ in the platformer environment producing interaction traces. These traces are then aggregated into signals that drive an optimization step on that morphology, yielding an updated morphology $e'$ . This interaction-optimization loop repeats for a fixed number of iterations, gradually transforming a starting morphology into a new one that better satisfies the gating criteria.

Figure \ref{fig:fullpipeline}  summarizes this process. The pipeline begins with an enemy morphology initialization step, which samples a fixed set of starting 4x4 morphologies from the space $E$. After this one-time initialization, the core of the system consists of a two- stage loop applied to one morphology at a time.  In the player control stage, agents with the base and augmented configurations act in the environment against a chosen morphology and generate interaction traces. In the morphology optimization stage, the traces collected for that morphology are used to produce an updated morphology. The loop then returns to the player control stage for the same morphology, and this two-stage process is repeated for a fixed number of iterations. The initial set is therefore only used as a collection of independent starting points.

We instantiate this general template in three variants differing in the morphology optimization component. In the Neural Enemy Field (NeEF) variant, the agents are controlled by reinforcement-learning policies, and morphology updates are produced by a neural generator trained on the resulting interaction traces. In the A*-based Production Rules variant, the agents follow plans from an A* path planner, and morphology updates are obtained based on following three simple production rules. In the RL-based Production Rules variant, the RL-controlled agents instead produce interaction data for the same production rules. In all experiments, the initial set of morphologies is identical across variants, ensuring that any observed performance differences arise from the choice of player control and morphology optimization strategy rather than from differing initial conditions.

\subsection{Augmented Configuration Mechanics}

As defined in Section 3, both agents always have access to a baseline movement set $B$ consisting of left and right horizontal movement and jump. On top of this baseline we consider a small set of extra mechanics obtained from prior work on automatic mechanic generation \cite{gonzalez2023mechanic}. The extra mechanics are as follows:

\begin{itemize}
    \item  Vertical teleport: instantly offset the player's position by a fixed vertical distance, moving the player up while preserving the horizontal position.
    \item  Horizontal teleport: instantly offset the player's position by a fixed horizontal distance, moving the player left or right while preserving the vertical position.
    \item Speed boost: multiplies the player's horizontal movement speed by a fixed factor, so that left and right movements are faster during the boost while jump dynamics remain unchanged.
    \item Double jump: allows the player to perform one additional jump while airborne before landing again, effectively adding a second jump that is only available once per airtime.
\end{itemize}

The speed boost mechanic was included as an edge case, as preliminary evaluation suggested it presents atypical gating challenges compared to the other mechanics.
The baseline configuration has access only to the baseline mechanics $B$, while the augmented configuration has access to $B \cup \{m\}$, where $m \in M$ and $M$ is the above set of new mechanics. All subsequent results compare these two configurations interacting with the same enemy morphology $e \in E$ to evaluate mechanic gating through morphology optimization.

\subsection{Agent Configuration Control}

In our experiments each agent is controlled in one of two different ways: by a Reinforcement Learning (RL) policy or by an A* path planner. In both cases, the available actions are determined by the configuration defined in Section 3. Under the baseline configuration the player can perform the three baseline mechanics, and under the augmented configuration the player can perform the same three mechanics plus the one extra mechanic $m$. The two approaches differ only in how they choose sequences of actions from these sets. 

In the RL implementation, the player is controlled using the Unity ML-Agents toolkit with Proximal Policy Optimization (PPO) as the learning algorithm. PPO is a widely used on-policy method that offers a good balance between stability and sample efficiency in discrete control problems\cite{schulman2017proximal,yu2022surprising}, and its integration in ML-Agents simplifies training directly within our Unity environment. 

The observation fed to the RL policy combines a low-dimensional vector with a ray-based perception module. The vector observation encodes a compact summary of the player’s action state (horizontal velocity, and vertical velocity, ). This provides precise, structured information that is inexpensive to process. 

The ray-cast perception module emits seven rays in a $180^\circ$ centred on the player, with a maximum length of 7 in-game units. Rays are configured to detect ground, lethal enemy cells, and weak enemy cells. The number of rays and their length were chosen to cover the entire 4×4 enemy morphology plus a small margin, while keeping the observation size modest. This gives the policy a local but informative view of nearby obstacles and vulnerable regions without resorting to high-dimensional image observations.  The action space is discrete and includes moving left, moving right, jumping, and when applicable invoking the extra mechanic $m$. This setup allows an RL agent both to navigate around the enemy morphology and to use the extra mechanic in context dependent ways.

The agent is trained to maximize the cumulative discounted reward over each episode. At every timestep $t$, the reward signal $R_t$ is define as the sum of a continuous step penalty and a terminal event reward:

\begin{equation}
    R_t = r_{\text{step}} + r_{\text{event}}
\end{equation}
The step penalty is applied at every timestep regardless of outcome:
\begin{equation}
    r_{\text{step}} = -\frac{3.5}{T_{\max}}
\end{equation}

Where $T_{max}$ is the maximum number of steps per episode. This small negative signal encourages the agent to resolve encounters efficiently rather than passively waiting. The terminal reward $r_{event}$ is issued once at the end of an episode and is determined by the outcome of the interaction:
\begin{equation}
    r_{\text{event}} =
    \begin{cases}
        +5.0 & \text{agent defeats enemy (weak cell)} \\
        -1.5 & \text{agent dies (lethal cell)} \\
        -5.0 & \text{agent exits the zone} \\
        0    & \text{no terminal event}
    \end{cases}
\end{equation}
The positive reward for defeating the enemy via a weak cell incentivizes the agent to actively seek and hit the enemy's vulnerable regions, while the penalties for death and zone exit discourage incorrect or evasive behaviour. Together, these signals shape an agent that reliably engages with the enemy morphology, providing meaningful interaction traces for the morphology optimization stage.

When our A* path planner controls an agent, we replace learning with search and treat enemy interaction as a deterministic planning problem over the same two types of sets of mechanics. 
For each cell in the 4×4 enemy grid, we run an A* search to determine whether there exists a plan that leads from the player’s random starting position to a collision with that specific cell. The heuristic is the Manhattan distance between the player’s current position and the target cell’s position, which is admissible under the grid-based geometry and inexpensive to compute. We perform these searches for both the baseline configuration and the augmented configuration  for each extra mechanic $m\in M$, yielding a reachability characterization of the entire morphology under both configurations. 
These reachability results form the interaction data for the A*-based Production Rules, and RL-based Production Rules pipelines: for each cell and configuration they indicate whether that cell can be reached and what kind of collision (weak or lethal) would occur. They are then used in the morphology optimization stage to identify and propose morphologies that satisfy the gating criteria. 

\subsection{Interaction Data}

For each enemy morphology and controller configuration, we generate interaction traces by running the Unity environment and logging relevant events to JSON. At the start of an episode, the enemy is spawned at a fixed position in the center of the scene. The player is spawned either to the left or to the right of the enemy, with this side chosen at random from two predefined positions.

In the RL case we train policies using the PPO configuration provided by ML-Agents. We modify some of the default values used on the configuration file, we use a batch size of 512, a buffer size of 2048, and five optimization epochs per update.
These values were based on initial experiments and were found to lead to good performance.
The policy and value networks share a simple multilayer architecture with four hidden layers of 128 units each. Training runs for 300,000 environment steps.

In the A* case we do not train a policy; instead, we perform up to sixteen search runs, one per tile in the 4x4 grid (fewer if some tiles are absent in the current morphology). Each search attempts to reach a particular tile, and we log whether that tile was successfully reached with both the baseline and augments controllers.

Each run, whether RL or A*, produces its own JSON file. For each enemy and mechanic, we have one file for the baseline controller and one for the augmented controller. These files contain aggregate statistics such as how many times the players defeated the enemy or died to it, as well as a time-ordered list of collision events. For each collision we record the local position of the enemy tile that was hit, the local position of the player at that moment, the direction of the player's movement, the type and index of the tile within the 4x4 grid, and whether the collision resulted in enemy destruction or player death. Together, these JSON files form the raw data that drive morphology generation.

\subsection{Dataset Construction}

The interaction traces are converted into cell-level training data for NeEF generator. Conceptually, we aim to learn from each mechanic $m$, how to transform an initial morphology into one that forces players to use $m$ to win. Operationally, we use a separate generator per mechanic and per pool of initial enemies, and we treat morphology generation as an iterative transformation process.  

Within each iteration, we next update the cell features. For each of the 16 cell in the grid we know it's current tile type, its coordinates within the morphology, and whether and how it was involved in collisions. We know whether the tile was ever reached, whether collisions with it led to enemy defeat or player death, and how often these events occurred under the baseline and augmented configurations. These statistics are aggregated into compact feature vectors. From them we derive labels indicating what tile type that position should take in the next iteration of the morphology one of our three generators, described below. Intuitively, positions that are important for enabling the extra mechanic (for example weak tiles that are only reachable when the mechanic is available) are encouraged, whereas tiles that allow the baseline agent to win are discouraged. 

All intermediate enemy morphologies contribute data in this way. We always start from the same ten randomly generated morphologies for these experiments, and for each iteration we produce a single updated morphology for each initial morphology. We do not discard morphologies in a genetic-algorithm sense; instead, the optimization signal comes from the collision statistics accumulated across iterations. When training the Neural Enemy Field (NeEF), we use a common validation split from the gathering data to monitor overfitting. In our experiments, evaluation is performed by re-running agents on the final generated enemies.

\subsection{Morphology Generators}

We explore three variants of the morphology generator that differ in how they use interaction data: (1) a  Neural Enemy Field (NeEF) driven by RL-collected traces, an A*-informed approach relying on Production Rules to define enemy morphology cell types, and an RL-informed version with the same production rules. These three approaches were chosen to span a spectrum from fully learned, parameterized updates (NeEF), through a purely search-based (A*) which is the mostly used on the video games area, to a hybrid that reuses the richer exploration patters of reinforcement learning while keeping the update step simple and interpretable. 
This allows us to compare how much benefit is gained from learning a parametric generator versus relying on explicit search and deterministic update rules.

Mechanic identity is encoded at the environment level rather than as an explicit input to a single shared  model, this in order to run parallel environments and speed up the process. Each mechanic $m$ is implemented in its own instantiation of the Unity environment, and we use a unique generator for each mechanic. From the perspective of the neural network, the mechanic is implicit in which model we query, not in an explicit mechanic embedding.

\subsubsection{Neural Enemy Field (NeEF)}

In this variant, the morphology generator is a small feed-forward neural network, which we refer to as the Neural Enemy Field (NeEf). NeEF operates at the cell level. For each of the 16 positions in the 4x4 grid, its input consists of the cell-level features described in Section 4.4: a one-hot encoding of the current cell type, the cell's coordinates within the grid, and the interaction traces (how often the cell was reached and whether collisions at that cell led to enemy defeat or player death under the baseline and augmented configurations).

Given an enemy morphology $e$, we apply NeEf to all sixteen cell positions, obtaining a 16x3 array of class probabilities (empty, weak, lethal). We then reconstruct a new morphology by taking, for each position, the most probable cell type and converting it back into a one-hot encoding. The resulting 16-element one-hot representation defines the enemy morphology in the next iteration.

NeEF is trained separately for each mechanic $m$. Each collision with any cell contributes one training sample, with its feature vector as input and the desired next-iteration tile type at that position as the target. Once the new enemy morphology is generated, the agents interact again with the new morphology to gather new data that will be use for the following iteration. The dataset construction that was previously mentioned it is only used on the NeEf generator given that is the only with a Neural Network.

\subsubsection{A*-based Production Rules variant}

In the A*-based Production Rules variant, we do not train a neural network or any other parametric model. Instead, we use A* search traces and then apply deterministic rules to map the interaction features to a new tile type. 
For each cell position $i$ in the $4 \times 4$ grid, we run A* twice: once with the baseline configuration and once with the augmented configuration. From each run we record a binary reachability indicator $r_i^{\text{base}}, r_i^{\text{me}} \in \{0,1\}$, where
\[
r_i^{\text{base}} =
\begin{cases}
1, & \text{if the agent can reach the goal} i,\\
0, & \text{otherwise,}
\end{cases}
\]
\\\[
r_i^{\text{me}} =
\begin{cases}
1, & \text{if the agent can reach the goal } i,\\
0, & \text{otherwise.}
\end{cases}
\]
Thus a value of $1$ indicates that there exists at least one successful path to the goal that visits cell $i$ for the corresponding agent, while $0$ means that no successful path uses that cell.

The morphology update is defined as a rule-based mapping applied independently to each cell. Let $\mathcal{T} = \{\text{empty}, \text{weak}, \text{lethal}\}$ be the set of tile types, and let
\[
\phi_m : \{0,1\} \times \{0,1\} \to \mathcal{T}
\]
denote the deterministic update function. In our instantiation, we use the following scheme: if the  player with the baseline configuration can reach the goal of cell $i$ it means that the player with the augmented configuration also can, as such the cell becomes \emph{lethal}; if the player with the augmented configuration can otherwise reach the goal of cell $i$, then the cell becomes a \emph{weak} tile; and if neither agent can reach the goal of cell $i$, then the cell can be  \emph{empty, or any }. This can be written compactly as
\[
\phi_m(r_i^{\text{base}}, r_i^{\text{me}}) =
\begin{cases}
\text{lethal}, & \text{if } r_i^{\text{base}} = 1 \text{ and } r_i^{\text{me}} = 1,\\[4pt]
\text{weak},   & \text{if } r_i^{\text{me}} = 1,\\[4pt]
\text{empty or any},  & \text{if } r_i^{\text{base}} = 0 \text{ and } r_i^{\text{me}} = 0.
\end{cases}
\]
The last case (neither agent can reach the cell) could be modified to keep such tiles \emph{lethal} instead of \emph{empty} if desired; the crucial aspect is that $\phi_m$ is a fixed, deterministic mapping from reachability patterns to tile types, and no learning or parameter optimization is involved.

\subsubsection{RL-based Production Rules variant}

This variant reuses the deterministic rules update of the A*-based Production Rules variant , but replaces search-based reachability with reachability estimated from RL interaction traces. The difference here is fundamentally that RL roughly approximates the difficulty of reaching a cell. If there is a path to a cell, A* will find it. However, RL will not necessarily be able to reach locations taking long sequences of specific actions.

As before, we consider an agent with a baseline configuration and an agent with augmented configuration trained by RL. During evaluation, we roll out each policy and use collision and success statistics from these episodes to construct the same binary reachability indicators $r_i^{\text{base}}$ and $r_i^{\text{me}}$ as in the A*-based classification rules update (cf.\ previous subsection), setting $r_i^{\text{base}} = 1$ (resp.\ $r_i^{\text{me}} = 1$) if at least one goal-reaching trajectory of the corresponding policy visits cell $i$, and $0$ otherwise.

Once these indicators have been estimated from RL rollouts, the morphology update proceeds exactly as before: for each cell $i$ we apply the fixed mapping $\phi_m(r_i^{\text{base}}, r_i^{\text{me}})$ defined in the A*-based variant to obtain the next tile type in $\{\text{empty}, \text{weak}, \text{lethal}\}$. Intuitively, this variant keeps the simple, interpretable update rule $\phi_m$, but grounds the reachability pattern in the stochastic behaviour of trained RL agents rather than in deterministic A* search. No additional training or parameter optimization is introduced; RL is used solely to provide the empirical reachability patterns that feed into the existing deterministic update.

While A* determines exact reachability through deterministic search, the RL-based Production Rules variant estimates reachability from the stochastic behaviour of trained RL agents. As noted earlier in this section, a cell that A* identifies as reachable may not be reached by an RL agent, as RL will not necessarily reach locations requiring long sequences of specific actions. Furthermore, A* may identify sequences that are theoretically reachable yet require precise, lengthy chains of actions that would be unreasonable to expect of human players. The two variants therefore answer different questions about the generated morphology, and are not methodologically redundant. 

\subsection{Technical Details}

All Unity environments are compiled in server or headless mode, as no rendering is necessary. Training runs on a Linux machine equipped with a Ryzen 9 5950X CPU, 64 GB of RAM, and an NVIDIA RTX 4090 GPU. RL training used the TensorFlow-based implementation bundled with ML-Agents; the morphology generators are implemented in Keras/TensorFlow in Python. This setup allows adversary training and morphology generation to proceed in a reasonably efficient loop without requiring human intervention during data collection, where the approx running time for the NeEF env was near 11 hours per run, the A* Based production rule was 3 sec per run and the RL-based Production Rules was near 11 hours. 

\section{Evaluation}

\begin{table*}[!t]
  \centering
  \caption{Average and Standar deviation values across all mechanics (DJ, S, TX, TY), best values bolded.}
  \label{tab:results}
  \begin{tabular}{lccccc}
    \toprule
    Method & SRAA ($\uparrow$) & SRBA ($\downarrow$) & SD ($\uparrow$) & DO ($\uparrow$) & CT ($\downarrow$) \\
    \midrule
    A*-based Production Rules & 0.62$\pm$0.32   & \textbf{0.00$\pm$0} & \textbf{0.42 $\pm$ 0.11} & 0.36$\pm$0.08 & \textbf{2.38 s} \\
    RL-based Production Rules & 0.70 $\pm$0.41 & 0.10$\pm$0.17 & 0.19 $\pm$0.17 & 0.64$\pm$0.04 & 36065 s \\
    NeEF  & 0.72$\pm$0.42 & 0.35$\pm$0.28 & 0.18$\pm$0.06 & \textbf{0.65$\pm$0.008} & 36062 s \\
    GA-DA    & 0.75$\pm$0.43 & 0.33$\pm$0.40 & 0.07$\pm$0.13 & 0.05$\pm$0.05 & 42672 s \\
    GA-A       & \textbf{1.00$\pm$0} & 0.83$\pm$0.28 & 0.08$\pm$0.14 & 0.05$\pm$0.08 & 21778 s \\
    \bottomrule
  \end{tabular}
\end{table*}

We now evaluate the enemy morphologies produced by our approaches along four axes that follow directly from our design goals: 
\begin{itemize}
    \item how well the generated enemy morphologies realize the intended mechanic-gating behaviour.
    \item the diversity of output morphologies from their starting points and
    \item from the other outputs of the same generator; and
    \item how expensive they are to compute.
\end{itemize}

The evaluation covers four mechanics from our augmented set \{double jump (DJ), speed boost (S), horizontal teleport (TX), and vertical teleport (TY)\} and compares the final enemy morphologies produced by each generator variant (the three interaction-driven approaches and two GA baselines) on a shared pool of initial morphologies. For each combination of method and mechanic, we obtain a set of final morphologies from the optimization loop and then evaluate those morphologies with the corresponding controllers. As described earlier, the interaction-driven methods are run for a fixed number of optimization iterations per mechanic and initial morphology; to ensure a fair comparison, the GA baselines are configured to run for the same number of runs and enemy morphologies evaluations, so that all approaches operate under comparable computation budgets and differ primarily in how they use the time that they have.

\subsection{Metrics}

From these evaluation runs we compute five scalar metrics that form the basis of our analysis: success rate of the augmented agent (SRAA), success rate of the base agent (SRBA), self-difference (SD), difference to original (DO), and computation time (CT). SRAA measures how often the generated enemies can be defeated when the player has access to the target mechanic, and is defined as the proportion of evaluation episodes in which the corresponding mechanic-augmented agent defeats a final enemy morphology. Symmetrically, SRBA is the success rate of the base configuration, in which the agent has only the baseline movement set. High SRAA indicates that enemies are actually defeatable with the intended mechanic, whereas low SRBA indicates that they are not easily or not possibly defeated without it; together, these two quantities capture the core gating behaviour we seek, namely enemies that are reliably defeatable with the extra mechanic but resistant to the baseline configuration.

The remaining metrics characterize how each method explores morphology space and at what cost. Self-difference (SD) measures diversity within a generator’s output morphologies by averaging a normalized tile-wise distance between $4 \times 4$ grids over the three tile types; higher SD corresponds to a broader variety of shapes. Difference to original (DO) applies the same distance to compare each final morphology to its own initial version, providing a proxy for how many structural changes the method performed. Computation time (CT) is the average wall-clock time required to generate final enemies across all four mechanics (DJ, S, TX, TY), reflecting practical cost. In our results table we therefore report SRAA ($\uparrow$), SRBA ($\downarrow$), SD ($\uparrow$), DO ($\uparrow$), and CT ($\downarrow$) to summarize how the different optimization strategies trade off gating performance, diversity, non-trivial morphology change, and computational expense.

\subsection{Baselines}

To contextualize our different approaches, we implement Genetic Algorithm (GA) baselines since evolutionary optimization is the closest methodological analogue to morphology generation in robotics and virtual creatures, which is the most relevant adjacent literature to our setting. Both GA baselines search directly over the same $4\times4$ morphology encoding (a $16\times3$ one-hot genome over \{empty, weak, lethal\}) and use the same evolutionary operators: uniform crossover at the cell level and per-cell mutation ($p=0.1$), where mutation always switches a cell to a different tile type. We initialize each evolutionary run from the same fixed pool of ten randomly selected starting morphologies, and we cap the number of expensive environment evaluations so the GA performs the same number of morphology evaluations/updates as the learned and rule-based approaches, ensuring a comparable computation budget. Notably the GA approaches still took the longest to run. 

To distinguish optimization for defeatability from optimization for gating, we include two GA variants. \textbf{GA-A} evaluates candidates using only the augmented-configuration agent, so fitness rewards morphologies that the augmented agent can reliably defeat. \textbf{GA-DA} evaluates candidates with both the baseline and augmented agents, and assigns fitness to favor gating behavior—high success for the augmented agent while discouraging success for the baseline agent—thereby making the evolutionary objective directly align with our design goal. This two-baseline setup clarifies whether any observed gap is due to evolution as an optimizer or due to the additional difficulty of satisfying a comparative, constraint-like objective across two configurations.

We now distinguish the two evolutionary objectives. Each GA variant uses a different fitness function. GA-A evaluates candidates using only the augmented-configuration agent, assigning fitness as the raw number of episodes in which the augmented agent successfully defeated the enemy morphology.
\begin{equation}
    \mathcal{F}_{\text{GA-A}}(e) = W_{\text{nm}}(e)
\end{equation}
Where $W_{nm}(e)$ is the total number of wins recorded by the augmented agent against morphology $e$. This objective rewards any morphology that the augmented agent can defeat, without penalizing morphologies that the baseline agent can also defeat, and therefore does not directly optimize for gating behaviour.

GA-DA evaluates candidates with both agent configurations, assigning fitness to favour the gating objectives high success for the augmented agent while discouraging success for the baseline agent. The fitness is defined as a normalized differential kill rate:
\begin{equation}
    \mathcal{F}_{\text{GA-DA}}(e) = 
    \frac{\rho_{\text{nm}}(e) - \rho_{\text{om}}(e)}
    {\max\!\left(\rho_{\text{nm}}(e) + \rho_{\text{om}}(e),\, 1\right)}
\end{equation}
where $\rho_{\text{nm}}$ and $\rho_{\text{om}}$ are the win rates of the augmented and baseline agents respectively defined as:
\begin{equation}
    \rho_{\text{nm}}(e) = \frac{W_{\text{nm}}(e)}{\max(N_{\text{nm}}, 1)}, \qquad
    \rho_{\text{om}}(e) = \frac{W_{\text{om}}(e)}{\max(N_{\text{om}}, 1)}
\end{equation}
where $W_{\text{nm}}(e)$ is the number of wins recorded by the baseline agent against morphology  $e$, and $N_{\text{nm}},N_{\text{om}}$ are the total number of episodes run for the augmented baseline configurations, respectively. The denominator normalizes the difference by the total activity of both agents in the range /[-1, 1/]. A value close to 1 indicates strong gating behaviour, the augmented agent wins reliably while the baseline agent does not.

\section{Results}

Table~\ref{tab:results} summarizes the behaviour of all methods across the four augmented set mechanics in terms of SRAA, SRBA, SD, DO, and CT. Overall, the interaction-driven approaches achieve the intended gating behaviour more consistently than the genetic baselines, for the full tables see Appendix \ref{app:per-mechanic}. 
The A* based approach obtains a moderate average SRAA of $0.62$ while keeping SRBA at $0$ for all mechanics, meaning that its enemy morphologies are never defeated by the base agent in our evaluations. The RL Rules and NeEF variants achieve slightly higher average SRAA values ($0.70$ and $0.72$, respectively), but at the cost of allowing some wins to the base agent: RL Rules has a low but non-zero average SRBA of $0.10$, and the NeEF leaks substantially more with an average SRBA of $0.35$, driven in particular by the horizontal teleport (TX) condition where the base agent succeeds in $0.8$ of episodes. In contrast, the GA-A and GA-DA baselines reach high or even perfect SRAA (average $1.0$ for GA-A and $0.75$ for GA-DA) but also high SRBA ($0.83$ and $0.33$, respectively), indicating that the enemies they produce are often beatable by both agents and therefore fail to realize the intended gating behaviour with the extra mechanic. 
Per-mechanic breakdowns show that gating is relatively easy to achieve for double jump (DJ) and vertical teleport (TY) for all three methods, whereas speed boost (S) is challenging for all interaction-driven variants (SRAA near zero) and simply ignored by the GA baselines, which tend to generate enemies that both agents can defeat see Appendix \ref{app:per-mechanic} for tables breaking down the per-metric results.

The speed boost (s) mechanic warrants special consideration. 
Unlike the other mechanics, the speed boost mechanic was included as a known edge case- in our prior work we found that it was generated due to an RL agent hacking the evaluation function \cite{gonzalez2023mechanic}.  
We included it as a known edge case, where we expected the morphology generation approaches to be unable to effectively produce a morphology with different possible collisions between a speed boost augmented player and a base player.  
The near-zero SRAA results therefore confirm an expected finding rather than representing a surprising failure: speed boost does not produce valid enemy morphologies for this setting, and its inclusion serves to identify the boundary conditions of the approach rather than demonstrate its typical performance.
In this case, because our approach is based on collision information, if a mechanic does not lead to unique collisions it will never be able to produce a satisfactory output for our morphology generation problem.

The per-mechanic breakdown in Table~\ref{tab:app_sr_base} and Table~\ref{tab:app_sr_desired} provides a more precise picture than the aggregate results in Table~\ref{tab:results}. Across DJ, TX, and TY, NeEF produces morphologies where the augmented configuration reliably defeats the enemy, with SRAA values of 0.9, 1.0 and 1.0. The standard deviations in Table~\ref{tab:results} are driven primarily by the speed boost mechanic, which, as discussed previously, was included as a known edge case. This affects all methods equally, and the high variance is therefore a property of the evaluation design rather than evidence of instability in any particular approach.

The morphology-level metrics highlight further differences between approaches. The A*-based Production Rules variant produces the most diverse final enemy sets, with the highest average self-difference ($\text{SD} = 0.42$), while also making non-trivial changes to its starting morphologies ($\text{DO} = 0.37$). The RL-based methods, particularly the NeEF, push morphologies even further away from their initial random shapes (average $\text{DO} \approx 0.64$--$0.66$) but do so in a more concentrated region of morphology space, with lower diversity ($\text{SD} \approx 0.18$--$0.20$) and some tendencies to collapse to similar solutions for a given mechanic. This follows from prior PCGRL work in terms of a tendency to converge to the same output \cite{earle2021learning,earle2024scaling}.

Both GA variants show very low DO (around $0.05$ on average), meaning that they rarely depart substantially from the initial random enemies, and their diversity is also limited (average $\text{SD}$ below $0.1$). Finally, computation time strongly distinguishes the methods: A*-based Production Rules variant completes in approximately $2.4$ seconds, whereas all RL and GA variants require on the order of tens of thousands of seconds, with GA~DA being the most expensive. 

Taken together, these results suggest that A*-based Production Rules variant offers the best overall trade-off between gating reliability, morphological diversity, and computational cost, the RL-based approaches provide more aggressive and flexible morphology transformations at the expense of some gating leakage and very high runtime, and the evolutionary baselines struggle to enforce gating despite substantial computational effort.
We anticipate that the relatively poor performance of evolution for this problem came from the lack of signal. 
All three other approaches are set up to use each collision as a signal for how to optimize the enemy morphology. 
Our GA baseline, as is typical of evolutionary approaches to morphology generation in the robotics field, instead only get a combined fitness signal after many collisions. 

\section{Visualization Case Study}

Our generators output enemy morphologies as a 4x4 grid of discrete collision cells, where each cell encodes one of the three interaction types described in Problem Definition. In this section we document a lightweight art workflow that treats each generated grid as inspiration that can be interpreted by an artist into enemy concepts with minimal additional context.

\begin{figure}
    \centering
    \includegraphics[width=1\linewidth]{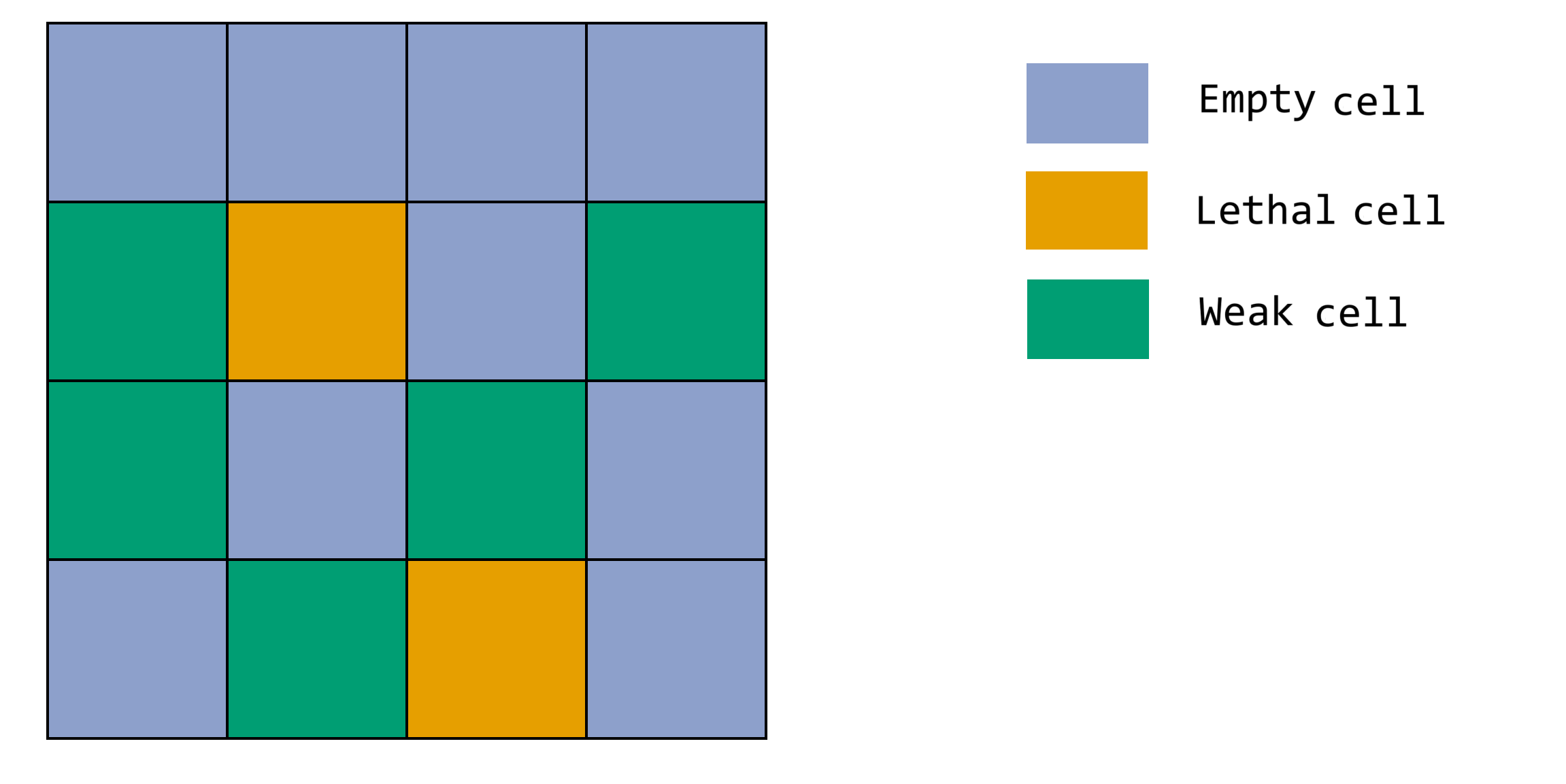}
    \caption{Enemy visualization}
    \label{fig:enemyvisual}
\end{figure}

To evaluate interpretability, we provided an artist with only the 4x4 grid produced by the generator, and the meaning of the three tile types (weak/lethal/empty) as seen in Figure \ref{fig:enemyvisual}. The artist's task was to create an enemy sprite concept that visually ``reads" as a coherent creature, and preserves the morphology's collision semantics by mapping visible body regions to the underlying tile types(e.g. exposed regions aligned with weak tiles; armored/spiked regions aligned with lethal tiles).
The artist used the morphology as an underlay and designed a silhouette that fits it's shape, while using shape language to communicate gameplay affordances. The resulting sprite was then overlaid on the original grid to verify alignment between the intended hit regions and the collision blueprint. Examples of this overlay are shown in Figure \ref{fig:teaser}, illustrating that the generator output can function directly as artist inspiration without requiring additional metadata from the optimization process. Across examples, we observed that even at a coarse 4×4 resolution, the grids support a practical range of readable silhouettes (compact bodies, overhangs, hollow centers, and asymmetric profiles. For more examples please see the Appendix\ref{app:visualization}.

Based on this case study, we argue that our enemy morphology generation can fit into the game development pipeline.
Generators can be used for rapid ideation, and mechanic-focused variation, designers can curate outputs using self-selection or simple performance criteria (Desired success vs Base success), and artists can treat the morphology as inspiration while freely exploring visual themes.

\section{Discussion}

Based on the results of the different approaches we acknowledge that the computational cost of the RL-based method is disproportionate to the problem complexity when compared to the A*-based variant. 
For the specific case of a 4x4 discrete grid, A* is strictly more efficient and produces exact solutions. 
However, according to some of the metrics in Table~\ref{tab:results}, the RL performance falls within the same distribution.
Thus claims about the relative A* and RL performance should be interpreted as reflecting a statistical tendency rather than a deterministic advantage.  
We include the RL approach as a proof of concept particularly useful for settings that lack an explicit transition model, a model that allows an agent to forward simulate to determine the impacts of its decisions, which A* requires.

A natural question is whether constraint satisfaction approaches such as Answer Set Programming (ASP) would be more appropriate baseline for this problem. We acknowledge that for purely structural constraints over a discrete 4x4 grid, such methods would likely be computationally superior. However, as demonstrated throughout this paper, the validity of a generated morphology cannot be determined by evaluating a fixed set of conditions over the grid alone. As formalized in Section 3, the intended gating behaviour is only observable through the win/loss behaviour produced when the baseline and augmented configurations interact with a given enemy morphology. A constraint encoding that captures this without relying on player-enemy interaction traces remains an open research challenge.

\section{Conclusions}

In this paper, we explore collision-based enemy morphology generation for 2D platformer-like enemies. We focused on the design goal of producing enemies that are defeatable with a specific augmented mechanics  while remaining resistant to a baseline moveset .We explored three different morphology generators, finding that all three could outperform baseline evolutionary approaches. 
We demonstrated, via a case study, the possibility of applying this approach in a production pipeline.



\begin{acks}

This work was funded by the Canada CIFAR AI Chairs Program. We acknowledge the support of the Alberta Machine Intelligence Institute (Amii). We acknowledge the support of the Natural Sciences and Engineering Research Council of Canada (NSERC). The authors would like to thank to Ana Karen Rodea for her awesome work with the enemy design.
\end{acks}

\bibliographystyle{ACM-Reference-Format}
\bibliography{mainbib}
\appendix

\begin{figure*}[!t]
    \centering
    \includegraphics[width=1\linewidth]{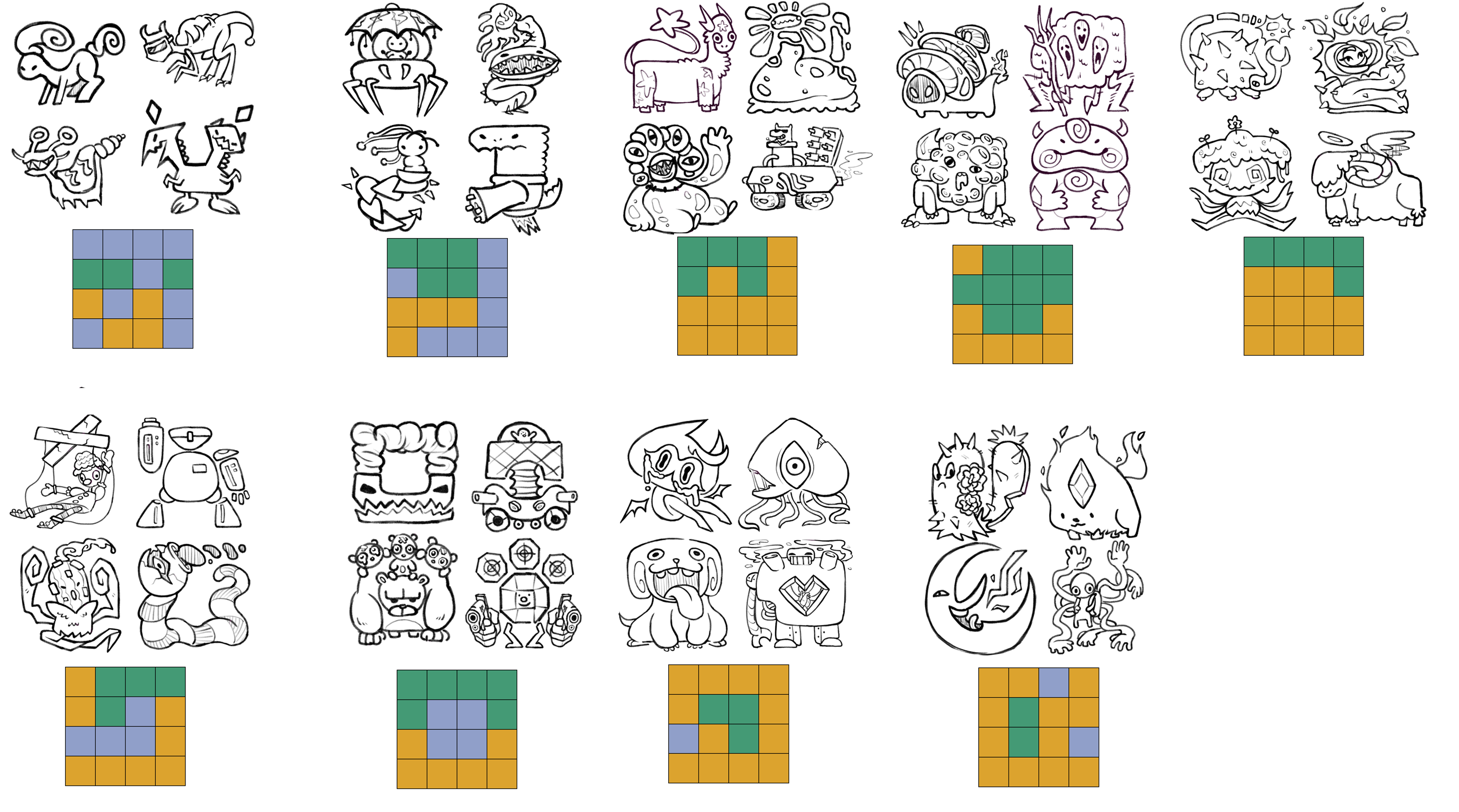}
    \caption{Randomly selected outputs that were delivered to an artist to conceptualize 4 different sprites per output. }
    \label{fig:fullart}
\end{figure*}

\section{Random Enemies Initialization Visualization}

This appendix reports the randomly generated initial enemy set used to initialize all experimental runs. The intent of including these examples is to make the starting conditions transparent and reproducible, and to provide visual context for interpreting how far each method transforms its initial morphology during optimization.

Each initial enemy is represented using the same 4×4 representation and the cell types employed throughout the paper. We do not include additional metadata (e.g., controller traces, fitness components, or method-specific parameters) is included in these visualizations; the figures are intended to be read strictly as collision layouts.

\begin{figure}
    \centering
    \includegraphics[width=\linewidth]{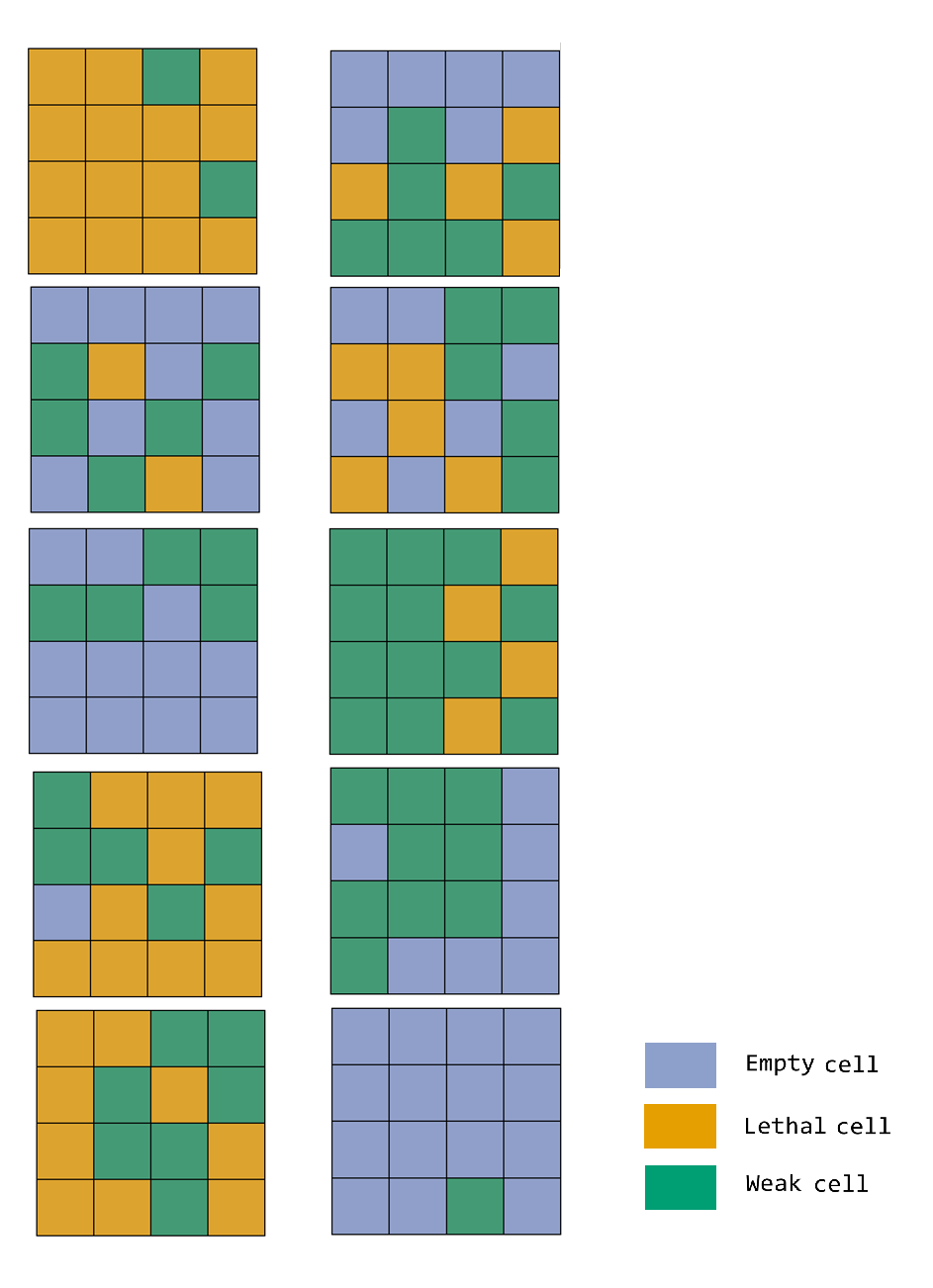}
    \caption{Visualization of the 10 random enemies initializations for all the experiments }
    \label{fig:enemies random}
\end{figure}

All methods (NeEF, A*-based Production Rules, RL-based Production Rules, and the evolutionary baselines) are seeded from this same fixed pool of initial enemies to ensure that performance differences are attributable to the generation approach rather than to differences in starting morphologies see Figure \ref{fig:enemies random}.

\label{app:per-mechanic}

\section{Per-mechanic Results}
\label{app:per-mechanic}

This appendix provides the full per-mechanic breakdown of all quantitative results reported in the main paper. While the main Results section emphasizes aggregate comparisons (using the mean across mechanics) to keep the narrative concise, method performance can vary substantially by mechanic. The tables in this appendix therefore report results separately for DJ, S, TX, and TY, enabling readers to identify mechanic-specific effects and reproduce the reported averages.

For each metric, rows correspond to the evaluated generation methods, columns correspond to mechanics, and the Avg column is the arithmetic mean across the four mechanics. 
Tables 2–5 report, respectively: Success Rate for the Augmented Agent (SRAA), Success Rate for the Base Agent (SRBA), Self-Difference, and Difference to Original. Readers interested in how individual mechanics drive the aggregate trends discussed in Section 6 should refer to these tables.

\begin{table}[t]
\centering
\small
\caption{Success Rate (Augmented Agent) by mechanic.}
\label{tab:app_sr_desired}
\begin{tabular}{lccccc}
\toprule
Method & DJ & S & TX & TY & Avg \\
\midrule
A*-based Production Rules  & 0.9 & 0.1 & 0.6 & 0.9 & 0.625 \\
RL-based Production Rules     & 1.0 & 0.0 & 0.8 & 1.0 & 0.700 \\
NeEF    & 0.9 & 0.0 & 1.0 & 1.0 & 0.725 \\
\addlinespace
GA-DA      & 1.0 & 1.0 & 1.0 & 0.0 & 0.750 \\
GA-A         & 1.0 & 1.0 & 1.0 & 1.0 & 1.000 \\
\bottomrule
\end{tabular}
\end{table}

\begin{table}[t]
\centering
\small
\caption{Success Rate (Base Agent) by mechanic.}
\label{tab:app_sr_base}
\begin{tabular}{lccccc}
\toprule
Method & DJ & S & TX & TY & Avg \\
\midrule
A*-based Production Rules & 0.0   & 0.0 & 0.0  & 0.0 & 0.0000 \\
RL-based Production Rules      & 0.4   & 0.0 & 0.0  & 0.0 & 0.1000 \\
NeEF     & 0.3   & 0.0 & 0.8  & 0.3 & 0.3500 \\
\addlinespace
GA-DA       & 0.0   & 1.0 & 0.33 & 0.0 & 0.3325 \\
GA-A          & 0.333 & 1.0 & 1.0  & 1.0 & 0.83325 \\
\bottomrule
\end{tabular}
\end{table}

\begin{table}[t]
\centering
\small
\caption{Self-Difference by mechanic.}
\label{tab:app_self_diff}
\begin{tabular}{lccccc}
\toprule
Method & DJ & S & TX & TY & Avg \\
\midrule
A*-based Production Rules & 0.54 & 0.29 & 0.32 & 0.54 & 0.4225 \\
RL-based Production Rules      & 0.50 & 0.15 & 0.06 & 0.08 & 0.1975 \\
NeEF     & 0.27 & 0.09 & 0.19 & 0.17 & 0.1800 \\
\addlinespace
GA-DA       & 0.00 & 0.00 & 0.31 & 0.00 & 0.0775 \\
GA-A          & 0.34 & 0.00 & 0.00 & 0.00 & 0.0850 \\
\bottomrule
\end{tabular}
\end{table}

\begin{table}[t]
\centering
\small
\caption{Difference to Original by mechanic.}
\label{tab:app_diff_original}
\begin{tabular}{lccccc}
\toprule
Method & DJ & S & TX & TY & Avg \\
\midrule
A*-based Production Rules& 0.31 & 0.47 & 0.43 & 0.26 & 0.3675 \\
RL-based Production Rules      & 0.57 & 0.66 & 0.68 & 0.66 & 0.6425 \\
NeEF     & 0.66 & 0.66 & 0.64 & 0.66 & 0.6550 \\
\addlinespace
GA-DA       & 0.00 & 0.00 & 0.10 & 0.10 & 0.0500 \\
GA-A          & 0.20 & 0.00 & 0.00 & 0.00 & 0.0500 \\
\bottomrule
\end{tabular}
\end{table}

\section{Visualization Case Study}
\label{app:visualization}

We include additional visualizations in this appendix, see Figure \ref{fig:fullart}. 
Across all the examples, we observed that even at a coarse 4×4 resolution, the grids support a practical range of readable silhouettes (compact bodies, overhangs, hollow centers, asymmetric profiles). Importantly, because the morphology encodes collision semantics rather than appearance, the same morphology can yield multiple distinct art outcomes depending on the visual theme, while still preserving the gameplay-critical weak/lethal regions. This suggests the representation is suitable for production pipelines where morphology is generated procedurally but the final appearance remains under human direction.

\end{document}